\documentclass{article}
\usepackage{spconf,amsmath,graphicx}
\usepackage{color}

\newcommand{\update}[1]{\textcolor{black}{#1}}

\title{Building a robust sentiment lexicon with (almost) no resource}
\name{Mickael Rouvier \qquad Benoit Favre}
\address{Aix-Marseille University \\
	CNRS, LIF UMR 7279 \\
	13000, Marseille, France}

\begin{document}

\maketitle

\begin{abstract}


Creating sentiment polarity lexicons is labor intensive. Automatically translating them from resourceful languages requires in-domain machine translation systems, which rely on large quantities of bi-texts. In this paper, we propose to replace machine translation by transferring words from the lexicon through word embeddings aligned across languages with a simple linear transform. The approach leads to no degradation,  compared to machine translation, when tested on sentiment polarity classification on tweets from four languages.


\end{abstract}

\section{Introduction}



Sentiment analysis is a task that aims at recognizing in text the opinion of the writer. It is often modeled as a classification problem which relies on features extracted from the text in order to feed a classifier. Relevant features proposed in the literature span from microblogging artifacts including hashtags, emoticons~\cite{go2009twitter,aoki2011method}, intensifiers like all-caps words and character repetitions~\cite{kouloumpis2011twitter}, sentiment-topic features~\cite{saif2012alleviating}, to the inclusion of polarity lexicons.

The objective of the work presented in this paper is the creation of sentiment polarity lexicons. They are word lists or phrase lists with positive and negative sentiment labels. Sentiment lexicons allow to increase the feature space with more relevant and generalizing characteristics of the input. Unfortunately, creating sentiment lexicons requires human expertise, is time consuming, and often results in limited coverage when dealing with new domains.

In the literature, it has been proposed to extend existing lexicons without supervision~\cite{andreevskaia2006mining,kanayama2006fully}, or to automatically translate existing lexicons from resourceful languages with statistical machine translation (SMT) systems~\cite{banea2008multilingual}. While the former requires seed lexicons, the later are very interesting because they can automate the process of generating sentiment lexicons without any human expertise. But automatically translating sentiment lexicons leads to two problems: (1) out-of-vocabulary words, such as mis-spellings, morphological variants and slang, cannot be translated, and (2) machine translation performance strongly depends on available training resources such as bi-texts.

In this paper, we propose to apply the method proposed in~\cite{mikolov2013exploiting} for automatically mapping word embeddings across languages and use them to translate sentiment lexicons
only given a small, general bilingual dictionary. After creating monolingual word embeddings in the source and target language, we train a linear transform on the bilingual dictionary and apply that transform to words for which we don't have a translation.


We perform experiments on 3-class polarity classification in tweets, and report results on four different languages: French, Italian, Spanish and German. Existing English sentiment lexicons are translated to the target languages through the proposed approach, given gs trained on the respective Wikipedia of each language. Then, a SVM-based classifier is fed with lexicon features, comparing machine translation with embedding transfer.


After presenting related work in Section~\ref{s:relatedwork}, the extraction of word gs and their mapping across languages are detailed in Section~\ref{s:approach}. The corpus on which experiments are carried out and the results of our experiments are presented in Section~\ref{s:experiments}. Finally, we conclude with a discussion of possible directions in Section~\ref{s:conclusions}.

\section{Related Work}
\label{s:relatedwork}

Many methods have been proposed for extending polarity lexicons: propagate polarity along thesaurus relations \cite{esuli2007pageranking,rao2009semi,hassan2014random} or use cooccurrence statistics to identify similar words~\cite{velikovich2010viability,kaji2007building}.

Porting lexicons to other languages has also been studied: use aligned thesauri and propagate at the sense level \cite{perez2012learning,gao2015cross}, translate the lexicon directly~\cite{hiroshi2004deeper,banea2013porting}, take advantage of off-the-shelf translation and include sample word context to get better translations~\cite{meng2012lost} or use crowd sourcing to quickly bootstrap lexicons in non-english languages~\cite{volkova2013exploring}.






\section{Approach}
\label{s:approach}



Our approach consists in creating distributional word representations in the source and target languages, and map them to each other with a linear transform trained given a small bilingual dictionary of frequent words. Then, source language words from the polarity lexicon can be projected
in the target language embedding. The closest words to the projecting are used as translation.


In our experiments, word embeddings are estimated on the source and target language Wikipedia corpora using the word2vec toolkit~\cite{mikolov2013efficient}. The embeddings are trained using skip-gram approach with a window of size 7 and 5 iterations. The dimension of the embeddings is fixed to 200.



\cite{levy2014neural} have shown that the skip-gram word embedding model is in fact a linear decomposition of the cooccurrence matrix. This decomposition is unique up to a linear transformation. Therefore, given two word representations created from the same cooccurrence matrix, a linear transform can be devised to map words from the first to the second. Assuming that cooccurrence matrices for the source and target languages are sampled from the same language-independent cooccurrent matrix, one can find a linear transform for mapping source words to target words, up to an error component which represents sampling error. This assumption is realistic for comparable corpora, such as embeddings trained on wikipedia in various languages. \update{In our experiments, we preferred to estimate word embeddings on Wikipedia rather than Twitter corpora because across languages, Tweets can cover different event from different countries, reducing the overlap.}

However, word embeddings represent a mixture from the senses of each word, making the cross-language mapping non bijective (a word can have multiple translations), which will probably contribute to the residual. Therefore, it should be reasonable to train a linear transform to map words between the source and target languages. Note that a linear transform would conserve the translations associated to linguistic regularities observed in the vector spaces.



The idea is to translate words in another language in the goal to generate sentiment lexicon. In~\cite{mikolov2013exploiting}, the authors propose to estimate a transformation matrix $W$ such that $Wx=y$, where $x$ is the embedding of a word in the source language and $y$ is the embedding of its translation in the target language. 

In order to estimate the $W$ matrix, suppose we are given a set of word pairs and their associated vector representations $\{x_{i}, y_{i}\}$ where $x_{i}$ is the embeddings of word $i$ in the source language and $y_{i}$ is the embedding of its translation. The matrix $W$ can be learned by the following optimization problem:
\begin{equation}
\min_{W} \sum_{i} \mid\mid Wx_{i} - y_{i} \mid\mid^{2}
\end{equation}

\noindent which we solve with the least square method.

At prediction time, for any given new word $x$, we can map it to the other language space by computing $y=Wx$. Then we find the words whose representations are closest to $y$ in the target language space using the cosine similarity as distance metric. In our experiments, we select all representations which cosine similarity is superior to $\lambda$ (with $\lambda=0.65$ set empirically).

In practice, we only have manual translations for a small subset of words, not necessarily polarity infused, on which we train $W$. We use that $W$ to find translations for all words of the sentiment lexicon.

\section{Experiments}
\label{s:experiments}


\begin{table*}[t]
\centering
    \begin{tabular}{ l | c c c c | c }
	\hline
	    & FR  &   IT & ES & DE & All \\
	\hline
    
    No Sentiment Lexicon              & 65.83 & 58.20 & 59.79 & 52.84 & 60.65 \\
    
    \hline
Pak et al, 2010 \cite{pak2010twitter}            & 65.97    & -    & - & -    & - \\
    Chen et al, 2014 \cite{chen2014building}    & 65.22    & 57.98 & 60.61 & 53.55 & 60.95 \footnotesize{\emph{\textbf{(+0.30 pt)}}}  \\
    Maks et al, 2014 \cite{MAKS14}       & 65.48 & 58.20 & 59.97 & -     & -\\

    \hline
    
    Moses (MPQA)                & 67.51 & 57.90 & 60.63 & 53.28 & 61.51 \footnotesize{\emph{\textbf{(+0.86 pt)}}} \\
    Moses (BingLiu)             & 67.48 & 58.13 & 60.99 & 52.81 & 61.70 \footnotesize{\emph{\textbf{(+1.05 pt)}}} \\
    Moses (HGI)                 & 66.47 & 57.58 & 60.49 & 53.91 & 61.16 \footnotesize{\emph{\textbf{(+0.51 pt)}}} \\
    Moses (NRC)                 & 66.98 & 58.27 & 60.70 & 54.80 & 61.56 \footnotesize{\emph{\textbf{(+0.91 pt)}}}\\

    \hline
    
    BWE (MPQA)                  & 67.38 & 58.35 & 60.61 & 53.24 & 61.53 \footnotesize{\emph{\textbf{(+0.88 pt)}}} \\
    BWE (BingLiu)               & 66.87 & 58.25 & 60.63 & 52.26 & 61.33 \footnotesize{\emph{\textbf{(+0.68 pt)}}} \\
    BWE (HGI)                   & 66.33 & 58.14 & 60.61 & 55.00 & 61.34 \footnotesize{\emph{\textbf{(+0.69 pt)}}} \\
    BWE (NRC)                   & 66.62 & 58.31 & 60.39 & 56.88 & 61.45 \footnotesize{\emph{\textbf{(+0.80 pt)}}}  \\

    \hline
    
    Moses + BWE (MPQA)          & 67.80 & 58.28 & 61.13 & 53.67 & 61.93 \footnotesize{\emph{\textbf{(+1.28 pt)}}} \\
    Moses + BWE (BingLiu)       & 67.77 & 58.76 & 61.00 & 54.07 & 61.95 \footnotesize{\emph{\textbf{(+1.30 pt)}}} \\
    Moses + BWE (HGI)           & 66.92 & 58.09 & 60.69 & 53.19 & 61.41 \footnotesize{\emph{\textbf{(+0.76 pt)}}} \\
    Moses + BWE (NRC)           & 66.73    & 58.42 & 61.02 & 55.23 & 61.72 \footnotesize{\emph{\textbf{(+1.07 pt)}}} \\
	\hline
	\end{tabular}
\caption{Results in macro-fmeasure obtained on the different languages (French, Italian, Spanish and German) using different sentiment lexicon (MPQA, BingLiu, HGI and NRC).}
\label{table:results}
\end{table*}

\subsection{Corpus and Metrics}

The sentiment polarity classification task is set as a three-class problem: positive, negative and neutral. The metrics used to measure performance is macro-fmeasure.
We developed our system on French and apply the same components on Italian, Spanish and German. A concise description of the training data follows.

The French (FR) corpus comes from the DEFT'15 evaluation campaign~\footnote{\url{https://deft.limsi.fr/2015/index.php}}. It consists of 7,836 tweets for training and 3,381 tweets for testing.
The Italian (IT) corpus was released as part of the SentiPOLC'14 evaluation campaign~\cite{basile2014overview}. It consists of 4,513 tweets for training and 1,930 tweets for testing.
For Spanish (ES), the TASS'15 corpus is used~\cite{tass2015}. Since the evaluation campaign was still ongoing at the time of writing, we use 3-fold validation on the training corpus composed of 7,219 tweets. 
German (DE) tweets come from the Multilingual Sentiment Dataset~\cite{narr2012language}. It consists of 844 tweets for training and 844 tweets for testing.







In order to extract features on those corpora, polarity lexicons are translated from English using the method described in Section~\ref{s:approach}.
The following lexicons are translated:

\begin{itemize}
\item \textbf{MPQA}: The MPQA (Multi-Perspective Question Answering) lexicon is composed of 4913 negatives words and 2718 positives words~\cite{wiebe2005annotating}.
\item \textbf{BingLiu}: This lexicon contains 2006 positive words and 4783 negative words. This lexicon includes mis-spellings, morphological variants and slang~\cite{hu2004mining}.
\item \textbf{HGI}: The Harvard General Inquirer (HGI) lexicons contains several dictionaries, we only used positive and negative lexicons that contains respectively 1915 and 2291 words~\cite{stone1968general}.
\item \textbf{NRC}: NRC Emotion Lexicon is a large word list constructed by Amazon Mechanical Turk~\cite{mohammad2013nrc}.
\end{itemize}

\subsection{System}

In order to test the value of the create lexicons, we use them in a typical sentiment polarity classification system~\cite{mohammad2013nrc_svm}. We first tokenize the tweets with a tokenizer based on macaon~\cite{nasr2011macaon}. Then, hashtags and usertags are mapped to generic tokens. Each tweet is represented with the following features and an SVM classifier with a linear kernel is trained to perform the task.

\begin{itemize}
\item Words n-grams
\item All-caps: the number of words with all characters in upper case
\item Hashtags: the number of hashtags
\item Lexicons: number of words present in each lexicon
\item Punctuation: the number of contiguous sequences of exclamation marks, question marks, and both exclamation and question marks
\item Last punctuation: whether the last token contains an exclamation or question mark
\item Emoticons: presence or absence of positive and negative emoticons at any position in the tweet
\item Last emoticon: whether the last token is a positive or negative emoticon
\item Elongated words: the number of words with one character repeated more than three times, for example : ``loooool"
\end{itemize}

We did not implement part-of-speech and cluster features as they cannot be assumed to be available in the target languages. \update{This system was part of the system combination that obtained the best results at the TASS 2015~\cite{tass2015,rouviertass2015} and DEFT 2015~\cite{hamon-EtAl:2015:DEFT,rouvier-favre-andiyakkalrajendran:2015:DEFT} evaluation campaigns.}

\subsection{Results}


Table~\ref{table:results} reports the results of the system and different baselines. The \emph{No Sentiment Lexicon} system does not have any lexicon feature. It obtains a macro-fmeasure of 60.65 on the four corpora. 

Systems denoted \cite{pak2010twitter}, \cite{chen2014building}, \cite{MAKS14} are baselines that correspond respectively to unsupervised, supervised and semi-supervised approaches for generating the lexicon. We observe that adding sentiment lexicons improves performance.

The \emph{Moses} system consists in translating the different sentiment lexicons with the Moses SMT toolkit. \update{It is trained on the Europarl bi-texts}. The approach based on translation obtains better results than the \emph{Baseline} systems. \update{In our experiments, we observe that some words have not been correctly translated (for example: slang words). The main drawback on this approach is that for correctly translating sentiment lexica, the SMT system must be trained on in-domain bi-texts.}.

\begin{figure}[htp] \centering{
\includegraphics[width=180px]{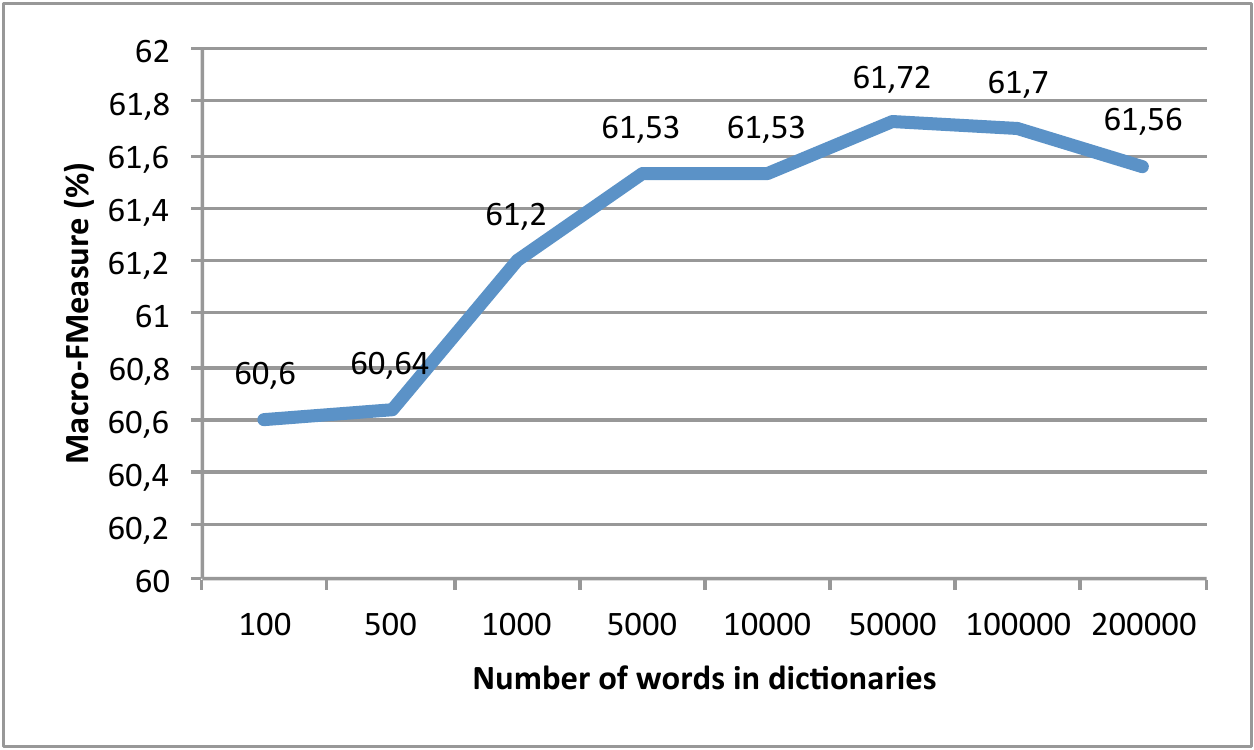}}
\caption{Average macro-fmeasure over the four languages when training the linear transform with a bilingual dictionary of n most frequent words (MPQA sentiment lexicon).}
\label{fig:dico_1}
\end{figure}

The \emph{BWE} (Bilingual Word Embeddings) system consists in translating the sentiment lexicons with our method. This approach obtains results comparable to the SMT approach. \update{The main advantage of this approach is to be able to generalize on words unknown to the SMT system.}

Moses and BWE can be combined by creating a lexicon from the union of the lexicons obtained by those systems. This combination yields even better results than translation or mapping alone.

Our second experiment consists in varying the size of the bilingual dictionary used to train $W$. Figure~\ref{fig:dico_1} shows the evolution of average macro f-measure (over the four languages) when the $n$ most frequent words from Wikipedia are part of the bilingual dictionary.  It can be observed that using the  50k most frequent words leads to the best performance (an average macro-fmeasure of 61.72) while only 1,000 words already brings nice improvements.

\begin{figure}[htp] \centering{
\includegraphics[width=180px]{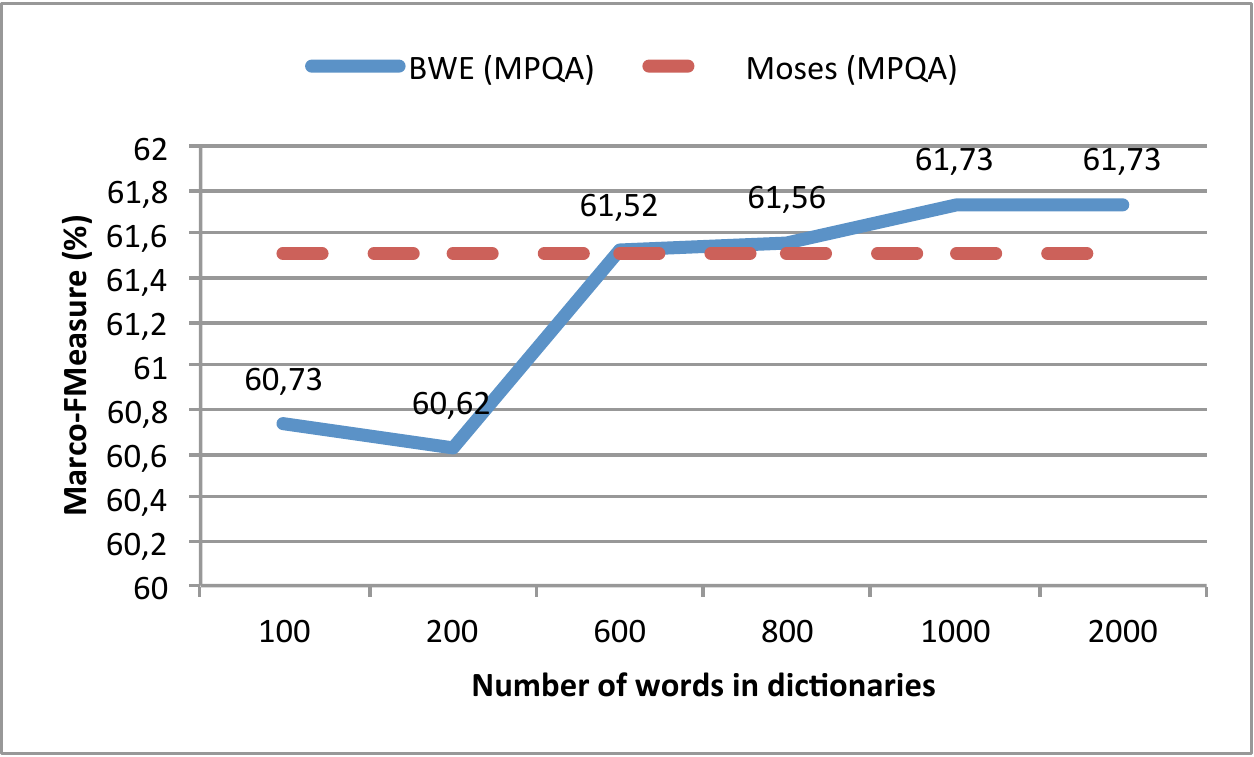}}
\caption{Average macro-fmeasure over the four language when training the linear transform with a small part of the lexicon words (MPQA sentiment lexicon).}
\label{fig:dico_2}
\end{figure}

In a last experiment, we look into the gains that can be obtained by manually translating a small part of the lexicon and use it as bilingual dictionary when training the transformation matrix. Figure~\ref{fig:dico_2} shows average macro-fmeasure on the four languages when translating up to 2,000 words from the MPQA lexicon (out of 8k). It can be observed that from 600 words on, performance is better than that of the statistical translation system.



\section{Conclusions}
\label{s:conclusions}

This paper is focused on translating sentiment polarity lexicons from a resourceful language through word embeddings mapped from the source to the target language. Experiments on four languages with mappings from English show that the approach performs as well as full-fledged SMT.
While the approach was successful for languages close to English where word-to-word translations are possible, it may not be as effective for languages where this assumption does not hold. We will explore this aspect for future work. 

\section*{Acknowledgments}

The research leading to these results has received funding from the European Union - Seventh Framework Programme (FP7/2007-2013) under grant agreement n\textsuperscript{o} 610916 SENSEI.

\bibliographystyle{IEEEbib}
\bibliography{acl2015}

\end{document}